\documentclass[conference]{IEEEtran}
\IEEEoverridecommandlockouts
\usepackage{cite}
\usepackage{svg}
\usepackage[numbers,sort&compress]{natbib}
\usepackage{multirow}
\usepackage{booktabs}
\usepackage{amsmath,amssymb,amsfonts}
\usepackage{algorithmic}
\usepackage{graphicx}
\usepackage{textcomp}
\usepackage{xcolor}
\def\BibTeX{{\rm B\kern-.05em{\sc i\kern-.025em b}\kern-.08em
    T\kern-.1667em\lower.7ex\hbox{E}\kern-.125emX}}
\begin{document}

\title{Part-Attention Based Model Make Occluded Person Re-Identification Stronger\\
}

\author{\IEEEauthorblockN{1\textsuperscript{st} Zhihao Chen}
\IEEEauthorblockA{\textit{Computer School} \\
\textit{Beijing Information Science and Technology University}\\
Beijing, China \\
2021011561@bistu.edu.cn}
\and
\IEEEauthorblockN{2\textsuperscript{nd} Yiyuan Ge$^{\ast}$ \thanks{*Corresponding author}}
\IEEEauthorblockA{\textit{School of Instrument Science and Opto-Electronics Engineering} \\
\textit{Beijing Information Science and Technology University}\\
Beijing, China \\
geyiyuan@bistu.edu.cn}
}

\maketitle

\begin{abstract}
The goal of occluded person re-identification  (ReID) is to retrieve specific pedestrians in occluded situations. However, occluded person ReID still suffers from background clutter and low-quality local feature representations, which limits model performance. In our research, we introduce a new framework called PAB-ReID, which is a novel ReID model incorporating part-attention mechanisms to tackle the aforementioned issues effectively. Firstly, we introduce the human parsing label to guide the generation of more accurate human part attention maps. In addition, we propose a fine-grained feature focuser for generating fine-grained human local feature representations while suppressing background interference. Moreover, We also design a part triplet loss to supervise the learning of human local features, which optimizes intra/inter-class distance. We conducted extensive experiments on specialized occlusion and regular ReID datasets, showcasing that our approach outperforms the existing state-of-the-art methods.
\end{abstract}


\section{Introduction}
Person re-identification (ReID) aims to retrieve specific pedestrians from different scenes and camera views and is of interest due to its wide range of applications in security \cite{b1}. Existing ReID methods learn by extracting a global representation of the target person. As shown in Figure 1(a), occlusion impacts the visual representation of pedestrians. When occlusion occurs, the distinction between various categories diminishes, leading to a situation where images with different IDs may have similar global representations. Also, as shown in Fig. 1(b), various occlusions widen the distance within the same category, which suggests that the global representations of the same pedestrian may be different. In summary, different parts of the occlusion and occluder contents can easily lead to wrong detection results.

There are two primary mainstream approaches for solving occluded ReID \cite{b50},\cite{b51},\cite{b52},\cite{b53},\cite{b54},\cite{b55},\cite{b56},\cite{b57},\cite{b58},\cite{b59},\cite{b60},\cite{b61},\cite{b62},
\cite{b63},\cite{b64},\cite{b65},\cite{b66},\cite{b67},\cite{b68},\cite{b69},\cite{b77},\cite{b78}, namely extra information-based methods and part-to-part matching methods. Among them, extra information-based methods use only the visible part of the body for ReID \cite{b2,b3,b4,b5}, which are usually based on extra information to judge the visibility of the body part (e.g., pose estimation). The above methods are computationally expensive and not robust when faced with complex occlusion situations (e.g., when occluding between pedestrians). The majority of the current approaches are based on part-to-part matching, which solves the occlusion problem by comparing the similarity between local features \cite{b38,b39,b40}. However, this part-to-part matching approach still faces some challenges: 1) The part-to-part matching-based approach relies on spatial attention maps \cite{b30} to construct body part features for ReID targets. However, the current ReID dataset lacks explicit annotations about the body part regions. 2) In occlusion scenes, the challenge remains to suppress background clutter and extract fine-grained effective features. 3) As shown in Fig. 1(c), the same body part may have a similar appearance across individuals. In other words, the visual appearance of body parts may not be inherently distinctive, making the conventional ReID loss, designed for learning global representations, less effective when applied to part representation learning.

\begin{figure}[tp]
\centerline{\includegraphics[width=0.45\textwidth]{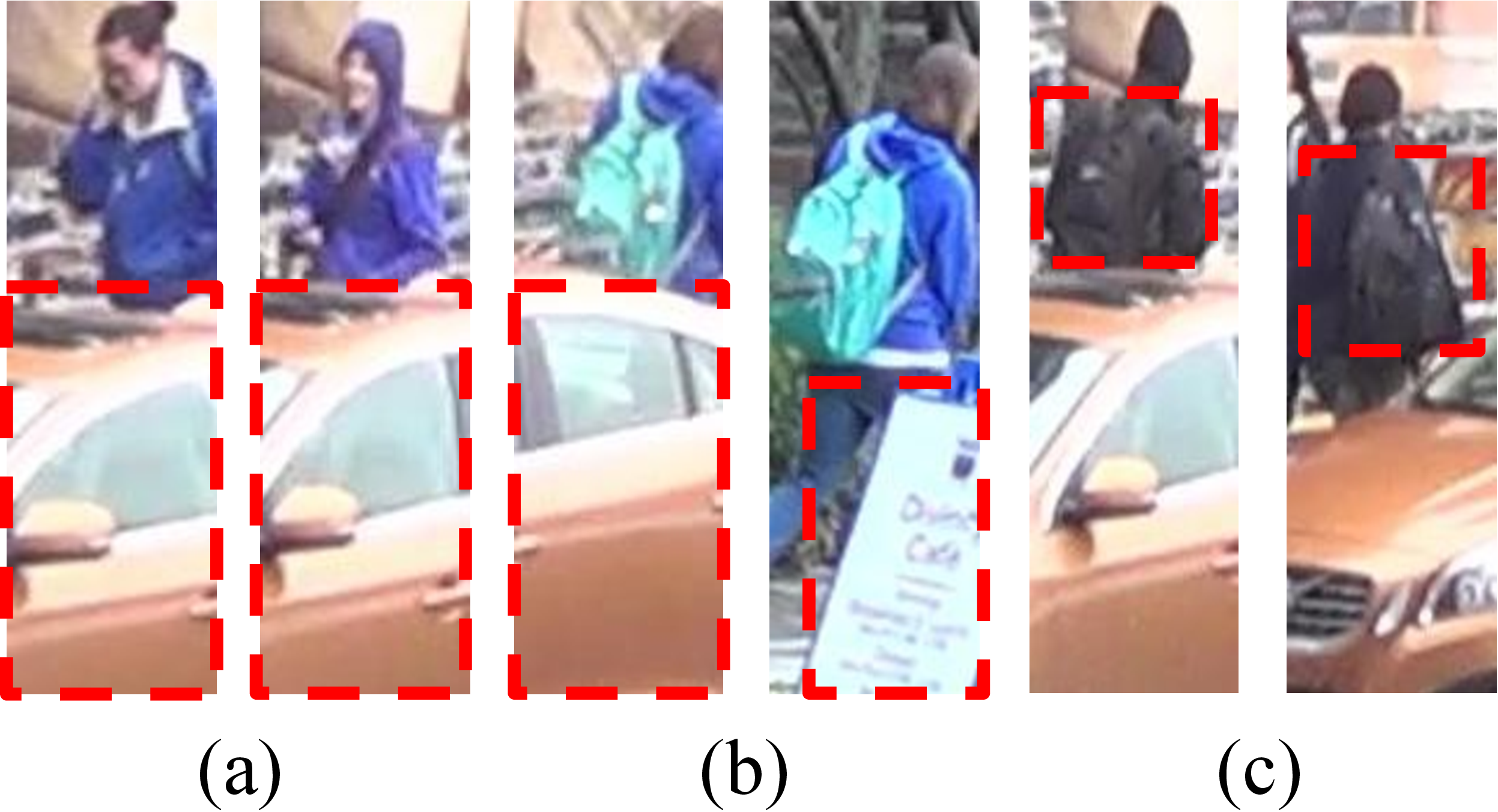}}
\caption{Examples of Challenges in Occluded Person Re-identification. In Figure. 1(a), similar occlusions reduce the gap between different categories. In Figure. 1(b), different occlusions can lead to an increase in distances within the same category. Figure. 1(c) illustrates that the same body part may have a similar appearance across different individuals.}
\label{fig}
\end{figure}

To solve the above problems, we propose the following solutions. \textbf{Firstly}, we introduce human parsing labels to guide the generation of attention maps and use pixel-level attention predictors to predict the attribution of each pixel to generate detailed attention maps. At the same time, dual-loss supervised training is used, with both body part prediction targets and ReID targets. This dual-supervision mechanism makes the final obtained attention map more relevant to the ReID task. \textbf{Secondly}, we then use a fine-grained feature focuser to process the generated attention maps. The effect of the attention map is enhanced by filtering irrelevant background information, and finally, fine-grained body part ReID features are generated. \textbf{Thirdly}, the standard ReID loss function assumes that different individuals have different appearances, i.e., different global feature vectors. The above assumption does not hold when confronted with part-based feature vectors. For this reason, we propose part triplet loss to supervise learning, which enhances the robustness of the model to similar part features.

Finally, we combine the part attention block with the global-local learning block to propose the part-attention based model PAB-ReID. Here are our contributions:
\begin{itemize}
\item We propose a part-attention based ReID model (PAB-ReID), which achieves person re-identification in the occlusion case by learning attention maps of different body parts.
\item We innovatively introduce the human parsing labels to guide the generation of the attention maps, which leads to more precise feature extraction regions for each body part.
\item We design a fine-grained feature focuser in a global-local learning block for part features, which can filter irrelevant background information and generate fine-grained body part features.
\item Part triplet loss is proposed for supervised learning of body part features, which is robust to similar body part features.
\end{itemize}

\section{Related Works}
\subsection{Person Re-Identification}
Person re-identification identifies and matches target pedestrians from existing video sequences, and most of the mainstream methods use CNN \cite{b70,b71,b72,b73,b74,b75,b76} architecture. The widespread effectiveness of convolutional neural networks in image processing has resulted in a growing adoption of neural network-based approaches for addressing ReID tasks, and the mainstream approaches can be divided into three categories. The first method uses local features to capture locally salient information about pedestrians by dividing the output of the CNN into several parts. Sun et al. \cite{b9} partially addressed the challenges posed by incomplete images and spatial misalignment. They achieved this by estimating the overlapping regions between two pedestrian images through the utilization of visibility-aware part-level features. The second approach is to represent pedestrians by global features of characters. Luo et al. \cite{b10} suggested various approaches to enhance the recognition efficacy of global features. Nonetheless, these methods that rely on global features exhibit limited performance in scenarios involving occlusion and alterations in the pedestrian's posture. The third approach is to combine local and global features, which are used to obtain a more significant representation of pedestrian features. Wang et al. \cite{b11} proposed Multigranular Network (MGN) in order to combine fine-grained local features with global features. Park et al. \cite{b12} introduced an RRID network that not only integrates global and local features but also leverages the interplay between different body parts. Combining a pre-trained backbone with a well-designed loss function is the popular pipeline, among which triplet loss \cite{b6} and cross-entropy loss \cite{b7} are the most widely used. To enhance the integration of cross-entropy loss and triplet loss, Luo et al. \cite{b8} introduced BNNeck as a designed mechanism.

\subsection{Occluded Person Re-Identification}
The ReID methods mentioned in references \cite{b6,b7,b8,b9,b10,b11,b12} assume the visibility of the entire pedestrian body, neglecting the more complex scenario of occlusion. In reality, pedestrians are frequently obscured by objects or other individuals. To address the aforementioned issues, a solution known as occluded person re-identification has been suggested. There are two main categories of mainstream occlusion re-identification methods, which are extra information-based methods and part-to-part matching methods.

The extra information-based approach uses pose estimation, segmentation, etc., to localize the human body parts. Wang et al. \cite{b16} used the poses estimation method to learn visible local features as well as topological information during training and testing phases to achieve high detection accuracy, which leads to higher computational overhead and slower inference. Miao et al. \cite{b41} employ pose landmarks to separate valuable information from occlusion noise. Despite the progress achieved by incorporating pose landmarks, the inability to train pose-guided region extraction and the constraints imposed by predefined landmarks visibility continue to hinder matching performance.

Part-to-part matching is an approach to solving the occlusion problem by comparing the similarity between local features. In their study, Zhang et al. \cite{b13} used an approach to achieve feature alignment in their study by finding the shortest paths between local features. Sun et al. \cite{b14} proposed a visibility-aware part model(VPM), which uses self-supervised learning to learn the visibility of a perceptual region. Jia et al. \cite{b15} introduced the Mask over Similarity (MoS) method in their study, which uses Jaccard similarity to measure the similarity between character images. Specifically, they reformulated the character recognition problem due to occlusion as an unaligned set-matching problem. However, this method always suffers from misalignment, missing parts, and background confusion.

In our work, the topological prior is introduced only in the training phase to restrict a more accurate local pooling region. In addition, we work on constructing more fine-grained and robust body part features through our designed focuser and part triplet loss.

\section{Methods}
\begin{figure*}[thp]
    \centering
    \includegraphics[width=1\textwidth]{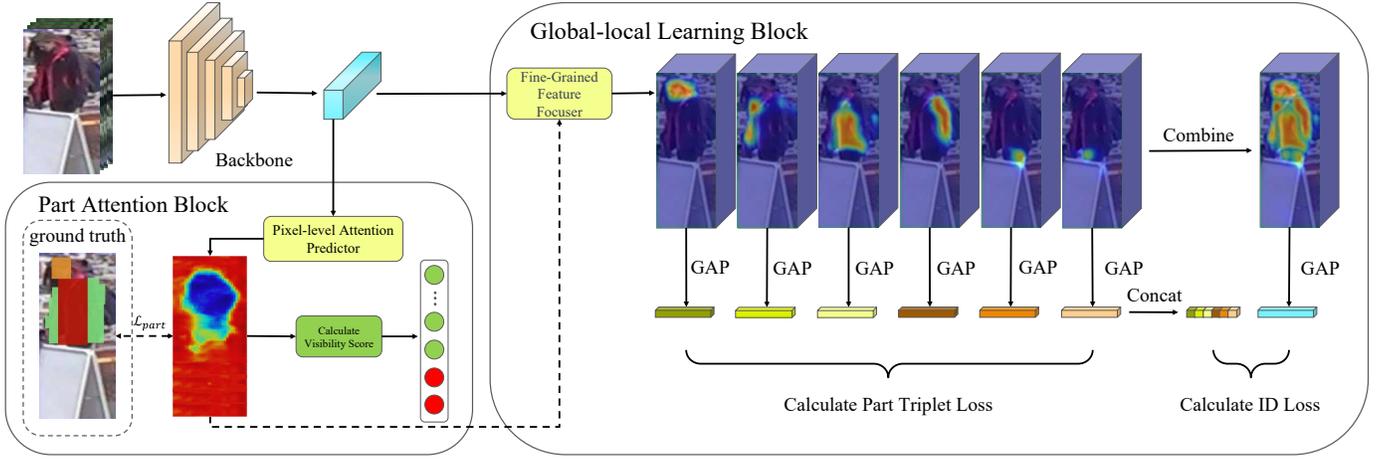}
    \caption{The detailed structure of  PAB-ReID is shown in the figure above. The PAB-ReID model comprises a part attention block responsible for generating part attention maps and a global-local learning block for extracting body part features. This paper generates six feature vectors representing the head, left hand, right hand, forehead, left leg, and right leg. In the above figure, the pedestrian's legs are occluded, and the visibility of the occluded leg is set to 0 when calculating the visibility score.}
    \label{fig}
\end{figure*}
The structure of our proposed part-attention based model is shown in Fig. 2. It consists of two core modules, the part attention block and the global-local learning block. In the part attention block, we introduce the human parsing labels to guide the generation of the part attention maps. Meanwhile, we filter irrelevant background information, and generate fine-grained body part features in the global-local learning block.
\subsection{Part Attention Block}\label{AA}
\begin{figure}[tp]
\centerline{\includegraphics[width=0.45\textwidth]{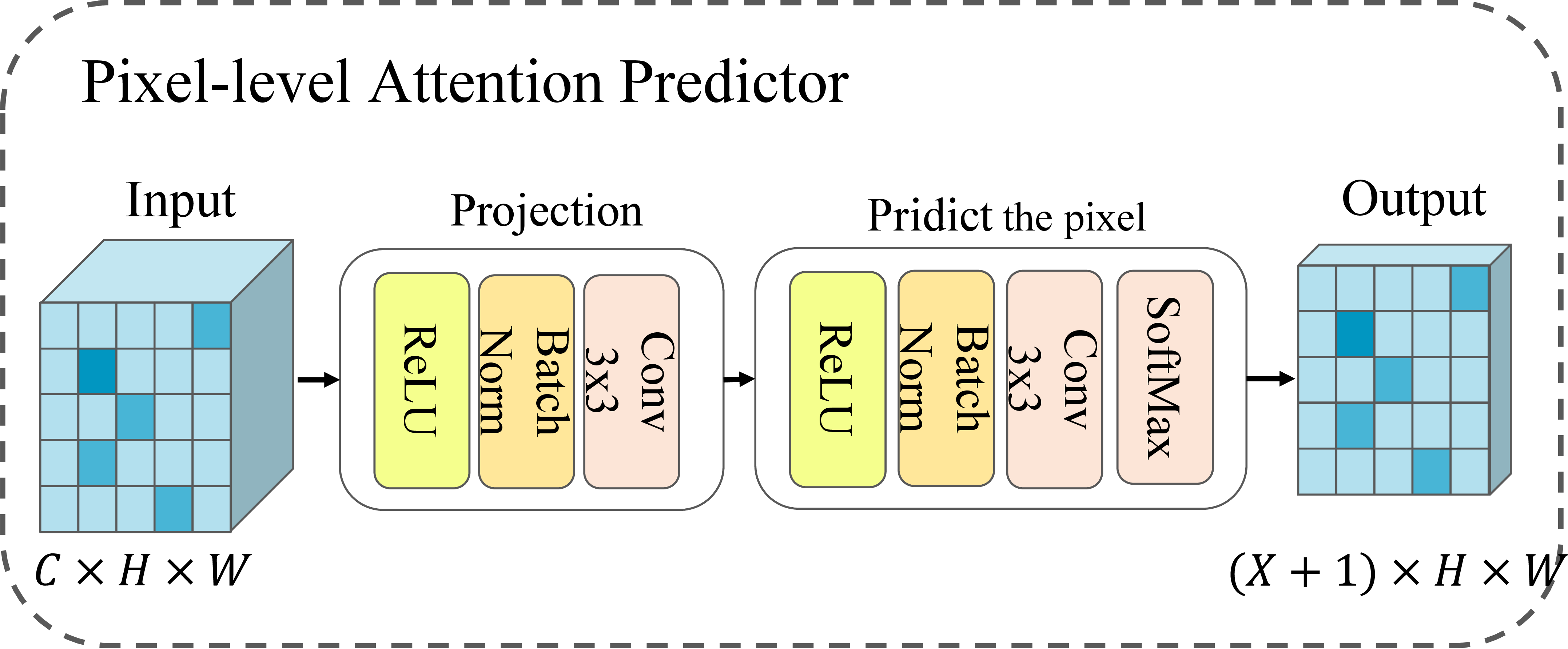}}
\caption{The detail architecture of pixel-level attention predictor.}
\label{fig}
\end{figure}
As shown in Fig.2,  the part attention block processes the features extracted by the backbone and generates a set of part attention maps that highlight specific body parts. In this block, we apply human parsing label to supervise the generation of attention maps for human body parts. This supervised labels consists of a number of roughly constrained regions, rather than pixel-level supervision. This is because the attention maps are also supervised by ReID-related losses, such as the ID loss and the Triplet loss, and thus the attention maps will focus on discriminative representation regions. In addition, we use part attention maps to compute the visibility of each body part. In the inference phase, we utilize only the visible body parts for the ReID task.
\paragraph{Pixel-level Attention predictor} The input feature map extracted by backbone is denoted as $B\in R^{C\times H\times W}$, which is subsequently processed by the pixel-level attention predictor. The process of the predictor is shown below: 
\begin{equation}
F_{1}=\operatorname{ReLU}\left(f\left(\delta_{3}(B)\right)\right)\label{eq1} \\
\vspace{-0.2cm}
\end{equation}
\begin{equation}
F=Softmax(ReLU(f(\delta_3(F_1))))\label{eq2}
\end{equation}
Where $\delta_{3}$ and $ReLU\left ( \cdot  \right ) $ denotes $3\times 3$ convolution operations and relu activation function, $f$ is the batch normalization, the softmax function is denoted as $Softmax\left ( \cdot  \right ) $. $F\in R^{\left ( x+1 \right ) \times H\times W} $ denotes $X$ part attention maps and a background attention map. In this paper, $X$ takes the value of $6$.
\paragraph{Part Attention Loss}We use part attention loss $\mathcal{L}_{part}$ to supervise the pixel-level attention predictor and the loss calculation process is as follows:
\begin{equation}
\mathcal{L}_{part}=-\sum_{x=0}^X\sum_{w=0}^{W-1}\sum_{h=0}^{H-1}\varepsilon\cdot\log\bigl(F_x(w,h)\bigr)\label{eq3} \\
\vspace{-0.2cm}
\end{equation}
\begin{align}
\varepsilon &= \begin{cases}
    \quad \frac{\theta}{N} & \text{where}\quad L\left ( w,h \right )=x \\
    1 - \frac{N - 1}{N} & \quad \quad \quad \text{others}
\end{cases}
\label{eq4}
\end{align}
\begin{figure}[tp]
\centerline{\includegraphics[width=0.45\textwidth]{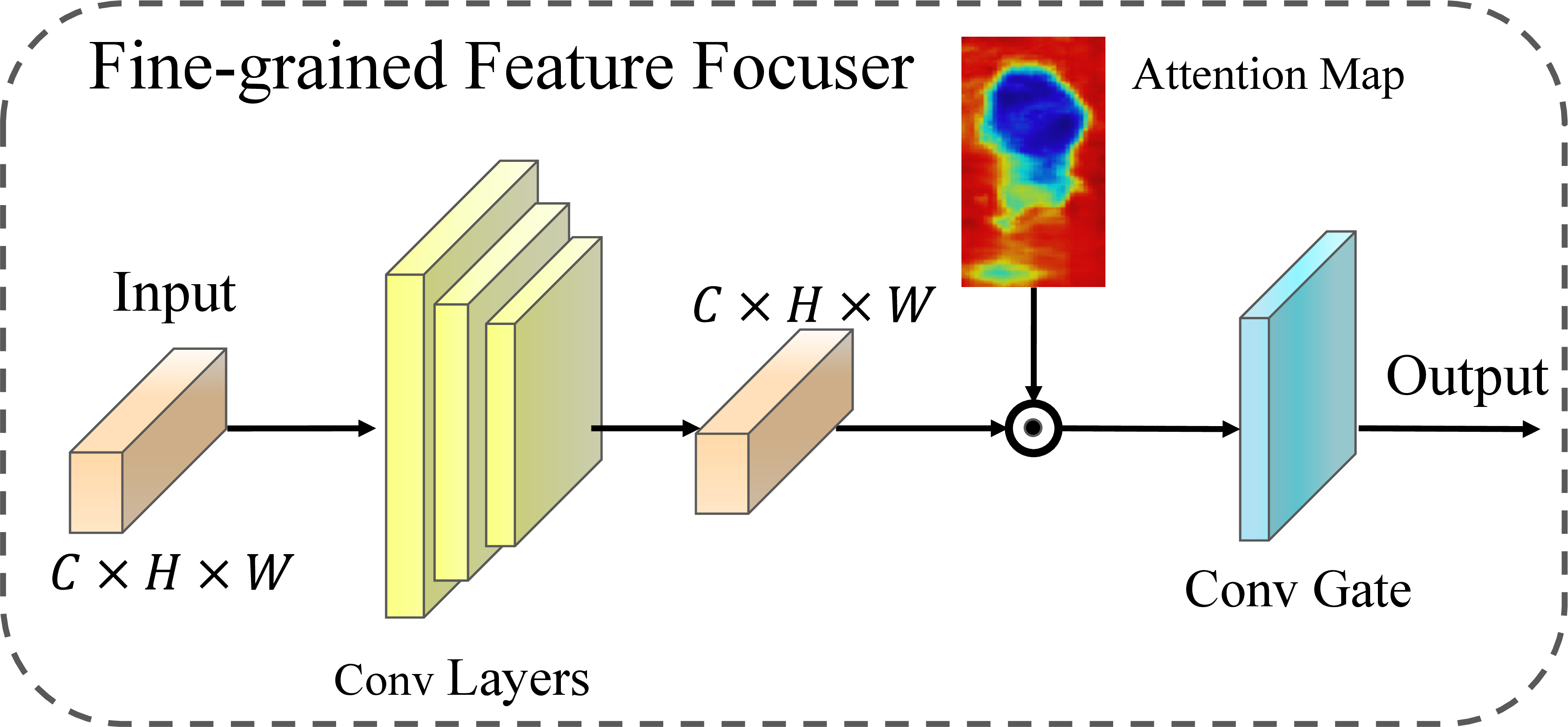}}
\caption{The detail architecture of fine-grained feature focuser is shown in the figure above.}
\label{fig}
\end{figure}Where $N$ is the batch size and $\theta $ is the label regularisation rate. If the pixel position $(w,h)$ of the human parsing label is attributed to the $x$-th body parts, the value of its pixel value $L(w,h)$ is set to $x$ and the background position is set to $0$.
\paragraph{Visibility Score}PAB-ReID computes a visibility score for each body part. We use $1$ to denote visible parts and $0$ to denote invisible parts. If at least one pixel value in a part attention map $F_x$, $x\in \left \{ 1,\cdots ,X \right \}$ is higher than the set value $\mu$ ($\mu$ is empirically set to $0.5$), the visibility score will be set to $1$.
\begin{align}
V_{s} &= \begin{cases}
    \quad 1&\text{where }max\left( F_{x}\left ( w,h \right )\right )>\mu \\
    \quad 0 & \quad \quad \quad \text{others}
\end{cases}
\label{eq5}
\end{align}
\subsection{Global-local Learning Block}\label{AA}
In the global-local learning block, our proposed fine-grained feature focuser applies part attention maps to the deep features extracted by backbone to obtain fine-grained body part features, after that gated convolution is introduced to filter the background clutter.
\paragraph{Fine-Grained Feature Focuser}The Pixel-level attention predictor generates $X$ attention maps highlighting the corresponding $X$ body parts. We first merge the $X$ body part attention maps to generate a foreground attention map $F_{f}\in R^{1\times H\times W} $:
\begin{equation}
F_{f}=Concat\left (F_{1},\dots, F_{x} \right)
\end{equation}
For the depth feature $B$ extracted by backbone, we further extract the feature representation of the whole body:
\begin{equation}
K_{1}=Conv\left (B \right)
\end{equation}
After that, we apply the B body part attention maps as well as the foreground attention map to the overall feature representation $K_1$:
\begin{equation}
P_{i}=K_{i}\odot F_{i},\quad i\in\left\{f,1\dots,X\right \} 
\end{equation}
$P_i$, $i\in\left\{f,1\dots,X\right \}$ denotes the fine-grained body part features and foreground feature. Moreover, we use gated convolution to filter the background contained in the features:
\begin{equation}
Q_{i}=Convgate\left(F_{i} \right),\quad i\in\left\{f,1\dots,X\right \} 
\end{equation}
\paragraph{$global-local$ ReID Loss}To increase the robustness of the model to non-discriminative features, we design the part triplet loss, and we denote the distance between two instances by the average distance of all body parts:
\begin{equation}
d_{parts}^{i,j}=\frac{1}{X}{\textstyle\sum_{x=1}^{X}}dist_{eucl}\left(f_{x}^{i},f_{x}^{j}\right)  
\end{equation}
Where $dist_{eucl}\left(f_{x}^{i},f_{x}^{j}\right)$ refers to the Euclidean distances of the two parts $f_{x}^{i}$,$f_{x}^{j}$, and subsequently, the losses are calculated using the average distances of the hardest positive and hardest negative parts $d_{parts}^{ap}$ and $d_{parts}^{an}$, respectively.
\begin{equation}
\mathcal{L}_{tri}\left(f_{0}^{a},\dots f_{X}^{a}\right )=\left [d_{parts}^{ap}-d_{parts}^{an}+\alpha\right ]+\\
\end{equation}
Where $\alpha$ is the part triplet loss margin, our proposed loss function makes it possible for model to focus on the most robust and discriminative parts during training, mitigating the effects of non-discriminative local and occluded features. The ablation experiments for the loss function in Fig. 5 show that part triplet loss is more effective than normal triplet loss.

In the ID classifier, our goal is to classify people with different identities, given instance $x_i$, with the identity label $D_i$. The loss function is calculated as follows:
\begin{equation}
\mathcal{L}_{ID}=-\sum_{i=1}^{N}log\left(\frac{y\left(x_{i},D_{i}\right )}{{\textstyle\sum_{j=1}^{N_{ID}}y\left(x_{i},D_{i}\right)}}\right )\\
\end{equation}
$N$ and $N_{ID}$ are the number of instances and body points, respectively, and $y\left(x_{i},D_{i}\right)$ denotes the probability of predicting $x_{i}$ as $D_{i}$.
\subsection{Training Procedure}\label{AA}The overall loss function to be optimized in the training phase is as follows:
\begin{equation}
\mathcal{L}_{sum}=\mathcal{L}_{tri}+\mathcal{L}_{ID}+\gamma_{part} \mathcal{L}_{part}\\
\end{equation}
Where $\mathcal{L}_{part}$ is the loss function used in the part attention block for supervised attention map generation, and $\gamma_{part}$ is used to control the contribution of $\mathcal{L}_{part}$ to the total loss, usually set to $0.35$.
\begin{table*}[htbp]
  \centering
  \caption{Comparison of methods on occluded datasets}
    \begin{tabular}{|c|c|c|c|c|c|c|c|}
    \hline
    \multirow{3}{*}{\textbf{Method}} & \multirow{3}{*}{\textbf{Venue}} & \multicolumn{6}{c|}{\textbf{Occluded Datasets}} \\
    \cline{3-8}
          &       & \multicolumn{2}{c|}{\textbf{Occluded-Duke}} & \multicolumn{2}{c|}{\textbf{Occluded-reID}} & \multicolumn{2}{c|}{\textbf{P-DukeMTMC}} \\
    \cline{3-8}
          &       & \textbf{Rank-1} & \textbf{mAP} & \textbf{Rank-1} & \textbf{mAP} & \textbf{Rank-1} & \textbf{mAP} \\
    \hline
    PVPM\cite{b33} & CVPR 20 & 47    & 37.7  & \textbf{-} & \textbf{-} & 85.1  & 69.9 \\
    \hline
    HG\cite{b34} & BMVC 21 & 61.4  & 50.5  & \textbf{-} & \textbf{-} & \textbf{-} & \textbf{-} \\
    \hline
    PAT\cite{b35} & CVPR 21 & 64.5  & 53.6  & 81.6  & 72.1  & -     & - \\
    \hline
    SSGR\cite{b36} & CVPR 21 & 69    & 57.2  & 78.5  & 72.9  & -     & - \\
    \hline
    OAMN\cite{b25} & CVPR 21 & 62.6  & 46.1  & -     & -     & -     & - \\
    \hline
    FED\cite{b27} & CVPR 22 & 68.1  & 56.4  & 86.3  & 79.3  & 83.1  & 80.5 \\
    \hline
    FRT\cite{b28} & TIP 22 & 70.7  & 61.3  & 80.4  & 71    & -     & - \\
    \hline
    RFCnet\cite{b29} & TPAMI 22 & 63.9  & 54.5  & -     & -     & 63.9  & 54.5 \\
    \hline
    HCGA\cite{b32} & TIP 23 & 70.2  & -     & 87.2  & -     & -     & - \\
    \hline
    BPBreID\cite{b30} & WACV 23 & 66.7  & 54.1  & 76.9  & 68.6  & 91    & 77.8 \\
    \hline
    CAAO\cite{b31} & TIP 23 & 68.5  & 59.5  & 87.1  & 83.4  & 92.5  & 81.4 \\
    \hline
    QPM\cite{b37} & TMM 23 & 66.7  & 53.3  & -     & -     & 90.7  & 75.3 \\
    \hline
    \textbf{Ours} &       & \textbf{72.6} & \textbf{63.5} & \textbf{87.4} & \textbf{87.1} & \textbf{93.1} & \textbf{83.2} \\
    \hline
    \end{tabular}%
  \label{tab:addlabel}%
\end{table*}%
\section{Experiments}
\subsection{Datasets and Settings}\label{AA}We conducted experiments on three mainstream occluded ReID datasets: Occluded-Duke \cite{b21}, Occluded-reID \cite{b22}, and P-DukeMTMC \cite{b22}. Occluded-Duke \cite{b21} is constructed based on the DukeMTMC \cite{b23} dataset. It comprises 15,618 training images for 702 identities, 2,210 occlusion query images for 519 identities, and 17,661 gallery images. There is a rich variety of variations in Occluded-Duke \cite{b21}, including different viewpoints and a wide variety of obstacles, including cars, bicycles, trees, and other people. The Occluded-reID \cite{b22} dataset was a new dataset captured by a mobile camera device containing 2,000 images of 200 people. For each identity, there are five full-body images of the person and five images of the person with different types of severe occlusions.P-DukeMTMC \cite{b22} is another subset of DukeMTMC \cite{b23}. This dataset includes 12,927 training images from 665 identities, 2,163 query images from 634 identities, and 9,053 gallery images. We also evaluate our model on the DukeMTMC-ReID \cite{b23} and Market-1501 \cite{b24} datasets. The DukeMTMC-ReID \cite{b23} dataset contains 36,411 images of 1,812 pedestrians. Of these, 1,404 pedestrians were captured by more than two cameras, while 408 pedestrians were captured by only one camera. The Market-1501 \cite{b24} dataset contains 32,668 images of 1,501 individuals from 6 cameras. We use rank-k (R@K) and mean average precision (mAP) for model evaluation.
\subsection{Implementations Details}\label{AA}We use Resnet-50 \cite{b20} after pre-training on ImageNet1K as the backbone network, and we remove the last down sampling of ResNet-50 in order to be able to extract deep features better. We use the Openpifpaf \cite{b42} and Mask R-CNN \cite{b43} to generate the human parsing labels. For data enhancement, images were first enhanced by random cropping and pixel filling, followed by random erasure with a probability of $0.5$. Each training batch consisted of $32$ samples of $8$ IDs with $2$ pictures per ID. Our model was trained in an end-to-end manner on $2$ NVIDIA RTX $3090$ GPUs for a total of $120$ epochs, using the Adam optimizer. The learning rate increases linearly from $3.5\times10^{-5}$ to $3.5\times10^{-4}$ after the first ten epochs and then decays to $3.5\times10^{-5}$ and $3.5\times10^{-6}$ at the $40$th and $70$th epochs, respectively. The marginal $\alpha$ of the 
ternary loss was set to $0.3$, and the label smoothing regularisation rate $e$ was set to $0.1$.
\subsection{Comparison with the State-of-the Art Method}\label{AA}As shown in Table 1, PAB-ReID exhibits advanced performance on the occluded dataset, achieving 72.6\%, 87.4\%, and 93.1\% Rank-1 and 63.5\%, 87.1\%, and 83.2\% mAP on Occluded-Duke, Occluded-reID, and P-DukeMTMC, respectively. compared with the cutting-edge methods in terms of Rank-1 by 1.9\%, 0.2\%, 0.6\% and mAP by 2.2\%, 3.7\%, 1.8\% respectively. In the Table 1, PAT \cite{b9} uses a local feature generation method, and due to the lack of a priori human topology information, the model is susceptible to interference from partial omissions and misalignments, leading to less effective results than PAB-ReID. Compared to the globally-based occlusion ReID approach HG \cite{b17}, PAB-ReID also demonstrates notable performance, showing an 11.2\% improvement in Rank-1 accuracy and a 13\% increase in mean average precision. This highlights the effectiveness of the partially-based method in addressing occlusion challenges, as the global method cannot achieve part-to-part matching. The Occluded-ReID dataset does not have a training set, our method still achieves excellent performance, which demonstrates that PAB-ReID possesses better domain adaptation.

We assess the performance of the models presented in this paper by comparing them with various cutting-edge ReID approaches, including FRT \cite{b20}, RFCnet \cite{b21}, HCGA \cite{b24}, BPBreID \cite{b22}, CAAO \cite{b23}, QPM \cite{b37}, etc., on the plain ReID and occluded ReID datasets, respectively, where we compare PAB-ReID with RFCnet \cite{b12}, BPBreID \cite{b22}, CAAO \cite{b23}, and HCGA \cite{b24} models on the common ReID dataset. As shown in Table 2, our proposed method achieves state-of-the-art performance on Market-1501 and DukeMTMC-ReID datasets. Specifically, our proposed method achieves 96.1\% Rank-1 and 89.5\% mAP on the Market-1501 dataset, which improves Rank-1 and mAP by 0.9\% and 1.1\%, respectively, compared to the existing state-of-the-art methods. Rank-1 of 91.2\% and mAP of 82.5\% were achieved on the DukeMTMC-ReID dataset. The above results show that PAB-ReID is also very effective when facing common person re-identification scenarios.
\begin{table}[htbp]
  \centering
  \caption{Comparison of PAB-ReID with the state of the art methods in normal ReID datasets}
    \begin{tabular}{|c|c|c|c|c|}
    \hline
    \multirow{3}{*}{Method} & \multicolumn{4}{c|}{\textbf{Normal Datasets}} \\
    \cline{2-5}          & \multicolumn{2}{c|}{\textbf{Market-1501}} & \multicolumn{2}{c|}{\textbf{DukeMTMC-ReID}} \\
    \cline{2-5}          & \textbf{Rank-1} & \textbf{mAP} & \textbf{Rank-1} & \textbf{mAP} \\
    \hline
    OAMN\cite{b25} & 93.2  & 79.8  & 86.3  & 72.6 \\
    \hline
    PAT\cite{b26} & 95.4  & 88    & 88.8  & 78.2 \\
    \hline
    FED\cite{b27} & 95    & 86.3  & 89.4  & 78 \\
    \hline
    FRT\cite{b28} & 95.5  & 88.1  & 90.5  & 81.7 \\
    \hline
    RFCnet\cite{b29} & 95.5  & 89.2  & 90.7  & 80.7 \\
    \hline
    BPBreID\cite{b30} & 95.1  & 87    & 89.6  & 78.3 \\
    \hline
    CAAO\cite{b31} & 95.3  & 88    & 89.8  & 80.9 \\
    \hline
    HCGA\cite{b32} & 95.2  & 88.4  & -     & - \\
    \hline
    \textbf{Ours} & \textbf{96.1} & \textbf{89.5} & \textbf{91.2} & \textbf{82.5} \\
    \hline
    \end{tabular}%
  \label{tab:addlabel}%
\end{table}%
\subsection{Ablation Studies}\label{AA}As shown in Table 3, we have taken ablation experiments for different modules in Occluded-Duke. The first line of Table 3 shows the performance when the part attention block is not included, the second line shows the performance when the fine-grained feature focuser is removed, the third line shows the performance when the pixel-level feature is not used predictor, the third line is the performance without using pixel-level attention predictor, and the fourth line is the ablation experiment using normal triplet loss.
\paragraph{Part Attention Block and Pixel-level Attention predictor}We conducted many ablation experiments for part attention block with pixel-level attention predictor to demonstrate the effectiveness of part attention block with pixel-level attention predictor. As shown in the first and third rows of Table 3, when removing the part attention block in the PAB-ReID, the Rank-1 of the model decreases by 7.2\%, and the mAP decreases by 7.3\%. It proves that using human parsing label to generate body part attention maps can better guide the model to learn the features of different body parts. When we remove the pixel-level attention predictor in the part attention block, the Rank-1 and mAP of the model decrease by
\begin{table}[htbp]
  \centering
  \caption{Ablation Experiments of PAB-ReID}
    \begin{tabular}{|l|c|c|}
    \hline
    \multicolumn{1}{|c|}{\multirow{2}[4]{*}{\textbf{Ablation Studies}}} & \multicolumn{2}{c|}{\textbf{Occluded-Duke}} \\
    \cline{2-3}
          & \textbf{Rank-1} & \textbf{mAP} \\
    \hline
    w/o Part attention block & 65.4  & 56.2 \\
    \hline
    w/o Fine-Grained Feature Focuser & 67.3  & 58.3 \\
    \hline
    w/o Pixel-level Attention predictor & 71.4  & 61.7 \\
    \hline
    w/o Part Triplet Loss & 67.1  & 57.1 \\
    \hline
    \textbf{ALL} & \textbf{72.6} & \textbf{63.5} \\
    \hline
    \end{tabular}%
  \label{tab:addlabel}%
\end{table}%
1.2\% and 1.8\%, respectively, indicating that the introduction of the pixel-level attention predictor enables
the model to generate more discriminative part attention maps.
\paragraph{Fine-Grained Feature Focuser}We conducted experiments on the Occluded-Duke dataset to demonstrate its effectiveness for fine-grained feature focuser. As shown in the third row of Table 3, the Rank-1 and mAP of the fine-grained feature focuser model decreased by 5.3\% and 5.2\%, respectively, when removed from PAB-ReID. The above results show that the fine-grained feature focuser learns fine-grained the body part and foreground information by mapping the partial attention to holistic features while suppressing background information through the gated convolution.
\paragraph{Part Triplet Loss}Part Triplet Loss guides the model to learn fine-grained body part ReID features. We conduct comparative experiments using normal id loss, triplet loss, and id + triplet loss on the Occluded-Duke dataset to showcase the efficacy of part triplet loss. As shown in Fig. 2, when using only triplet loss on top of PAB-ReID, Rank-1, and mAP are 67.1\% and 57.1\%, respectively, decreased by 5.5\% and 6.4\% compared to using part triplet loss. The above results show that part triplet loss optimizes the intra/inter-class distance and enhances the learning of robust features.
\begin{figure}[htbp]
\centerline{\includegraphics[width=0.45\textwidth]{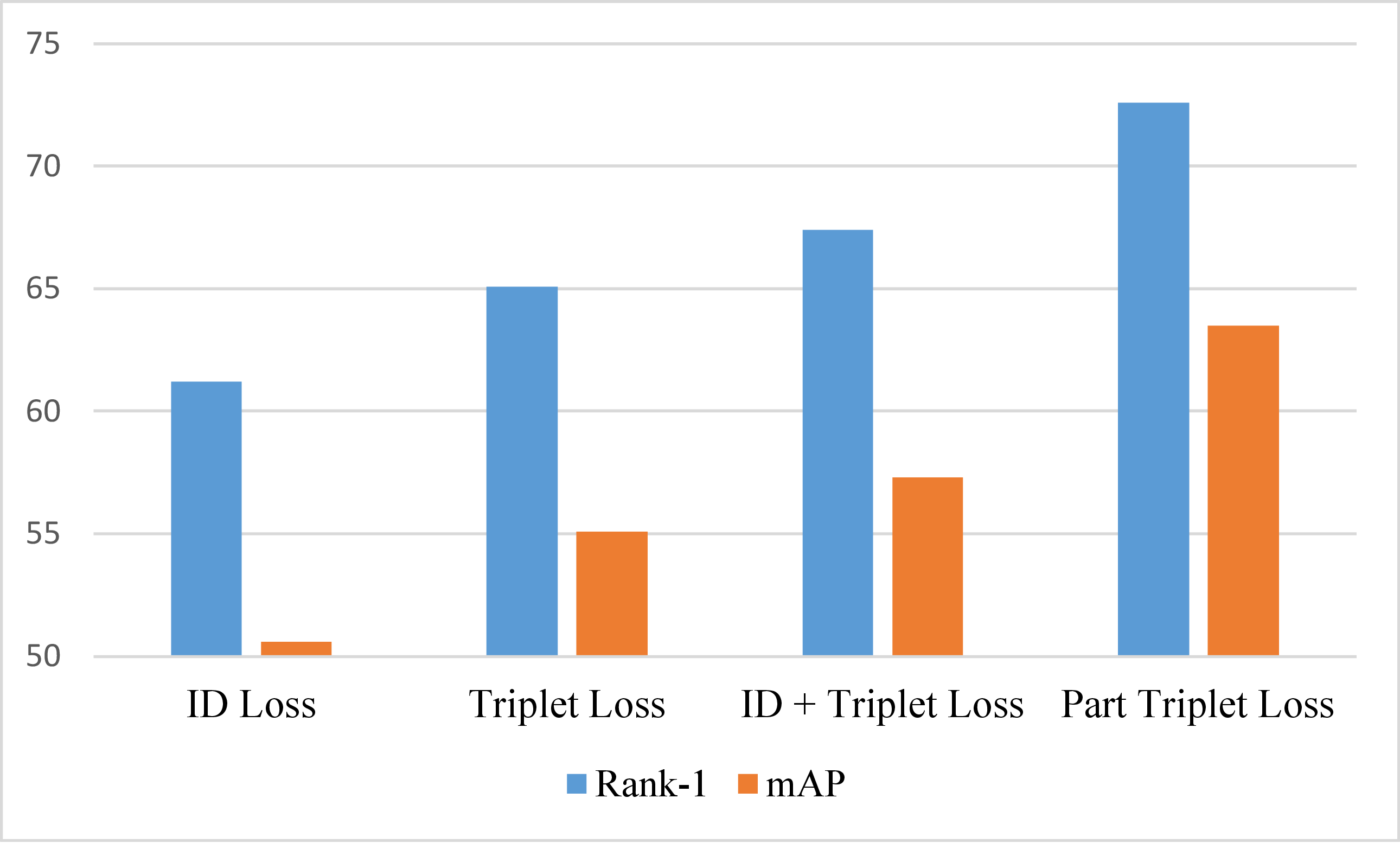}}
\caption{Ablation experiments of loss function. The first column represents ID loss, the second column represents triplet loss, the third column represents the sum of ID and triplet loss, and the fourth column represents part triplet loss. From the above figure, it can be observed that part triplet loss can better guide the model learning.}
\label{fig}
\end{figure}
\paragraph{Hyper-Parameter Sensitivity Experiments}We introduced two important hyperparameters $\mu$ and $\gamma_{part}$ in PAB-ReID, the combination of which determines the model's performance. In order to explore the optimal combination of $\mu$ and $\gamma_{part}$, we conducted an ablation experiment of the hyperparameters on the Occluded-Duke dataset, and the results of the experiment are shown in Fig. 6. We first set $\mu$ and $\gamma_{part}$ to $0.15$ and $0.3$, and then increased them by $0.2$, respectively. The experimental results show that when the visibility score is larger than $0.7$, the model takes part of the visible features as occluded features, which reduces the model performance and is used to regulate the proportion of $\mathcal{L}_{part}$ in the $\mathcal{L}_{sum}$. When $\gamma_{part}$=$0.55$, the proportion of $\mathcal{L}_{part}$ is larger, the model pays more attention to the generation of the attention graph in the part attention block and reduces the attention to the ReID task, resulting in a decrease in the Rank-1 of the model.
\begin{figure}[htbp]
\centerline{\includegraphics[width=0.45\textwidth]{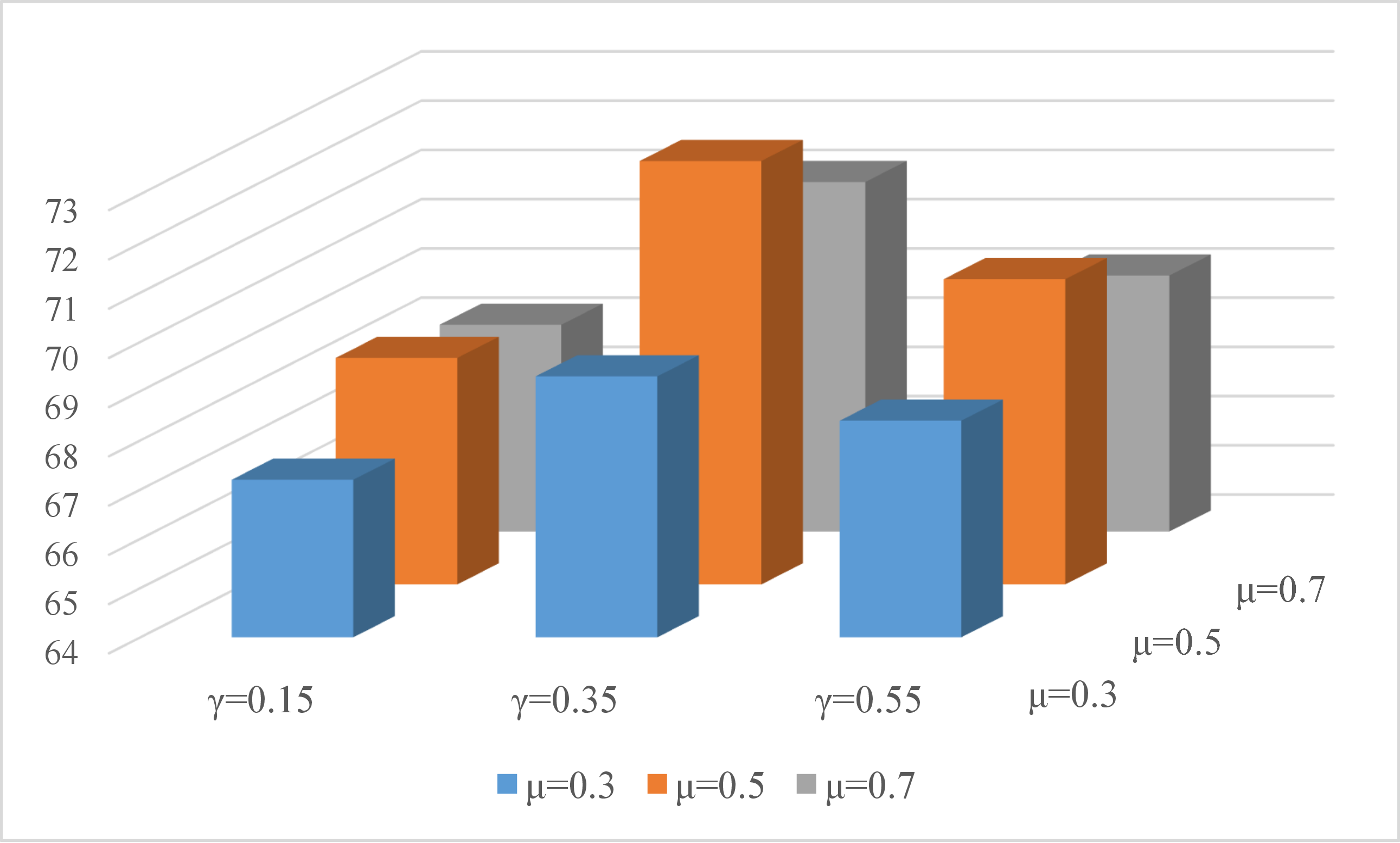}}
\caption{The Rank-1 results from hyperparameter sensitivity experiments. The model performs best when $\mu$ and $\gamma_{part}$ are set to 0.5 and 0.35, respectively.}
\label{fig}
\end{figure}
\section{Conclusion}
In this paper, we propose a novel part-attention based model (PAB-ReID) to address the challenges in occluded person re-identification. Firstly, we design the part attention block which utilize the external human semantic information to generate ReID-related body part attention maps.This part attention maps provide more accurate feature extraction regions for human body parts. Secondly, we also propose a fine-grained feature focuser for obtaining more granular pedestrian features while suppressing the background information. Thirdly, we proposed the part triplet loss to supervise the learning of body part feature, which enhances the robustness of the model to similar body part appearance. Finally, experiments on five rigorous datasets surface our method outperforming existing state-of-the-art methods.

\section*{Acknowledgment}
\begin{itemize}
\item Chen would like to thank Ge for being his guiding light in his research.
\item Chen would like to thank his parents and friends for supporting his scientific endeavors.
\item Supported by Promoting the Classification and Development of Colleges and Universities-Student Innovation and Entrepreneurship Training Programme Project-School of Computer (5112410852).
\end{itemize}

\vspace{12pt}

\end{document}